\documentclass[sigconf]{acmart}

\usepackage{booktabs} 
\DeclareMathOperator*{\argmax}{arg\,max}
\DeclareMathOperator*{\argmin}{arg\,min}

\copyrightyear{2020} 
\acmYear{2020} 
\setcopyright{licensedusgovmixed}
\acmConference[GECCO '20]{Genetic and Evolutionary Computation Conference}{July 8--12, 2020}{Cancun, Mexico}
\acmBooktitle{Genetic and Evolutionary Computation Conference (GECCO '20), July 8--12, 2020, Cancun, Mexico}
\acmPrice{15.00}
\acmDOI{10.1145/3377930.3390214}
\acmISBN{978-1-4503-7128-5/20/07}

\begin{document}

\title{Evolving Inborn Knowledge For Fast Adaptation in Dynamic POMDP Problems}
\subtitle{}

\author{Eseoghene Ben-Iwhiwhu}
\affiliation{%
  \institution{Department of Computer Science, Loughborough University, Loughborough, UK}
}
\email{e.ben-iwhiwhu@lboro.ac.uk}

\author{Pawel Ladosz}
\affiliation{%
 \institution{Department of Computer Science, Loughborough University, Loughborough, UK}
}
\email{p.ladosz2@lboro.ac.uk}

\author{Jeffery Dick}
\affiliation{%
 \institution{Department of Computer Science, Loughborough University, Loughborough, UK}
}
\email{j.dick@lboro.ac.uk}

\author{Wen-Hua Chen}
\affiliation{%
 \institution{Department of Aeronautical and Automotive Engineering, Loughborough University, Loughborough, UK}
}
\email{w.chen@lboro.ac.uk}

\author{\break Praveen Pilly}
\affiliation{HRL  Laboratories, California, USA}
\email{pkpilly@hrl.com}

\author{Andrea Soltoggio}
\affiliation{%
 \institution{Department of Computer Science, Loughborough University, Loughborough, UK}
}
\email{a.soltoggio@lboro.ac.uk}
\renewcommand{\shortauthors}{E. Ben-Iwhiwhu et al.}

\begin{abstract}
Rapid online adaptation to changing tasks is an important problem in machine learning and, recently, a focus of meta-reinforcement learning. However, reinforcement learning (RL) algorithms struggle in POMDP environments because the state of the system, essential in a RL framework, is not always visible. Additionally, hand-designed meta-RL architectures may not include suitable computational structures for specific learning problems. The evolution of online learning mechanisms, on the contrary, has the ability to incorporate learning strategies into an agent that can (i) evolve memory when required and (ii) optimize adaptation speed to specific online learning problems. In this paper, we exploit the highly adaptive nature of neuromodulated neural networks to evolve a controller that uses the latent space of an autoencoder in a POMDP. The analysis of the evolved networks reveals the ability of the proposed algorithm to acquire inborn knowledge in a variety of aspects such as the detection of cues that reveal implicit rewards, and the ability to evolve location neurons that help with navigation. The integration of inborn knowledge and online plasticity enabled fast adaptation and better performance in comparison to some non-evolutionary meta-reinforcement learning algorithms. The algorithm proved also to succeed in the 3D gaming environment Malmo Minecraft.
\end{abstract}

%
%

\ccsdesc[500]{Computing methodologies~Lifelong machine learning}
\ccsdesc[500]{Computing methodologies~Neural networks}
\ccsdesc[300]{Computing methodologies~Online learning settings}
\ccsdesc[300]{Computing methodologies~Partially-observable Markov decision processes}
\ccsdesc[300]{Computing methodologies~Genetic algorithms}

\keywords{lifelong learning, adaptive agent, self modifying network, neuroevolution, neuromodulation, few-shots learning, Hebbian learning}

\maketitle

\section{Introduction}
The field of deep reinforcement learning (RL) has showcased amazing results in recent time, solving tasks in robotic control \cite{duan2016benchmarking, lillicrap2015continuous}, games \cite{mnih2015human} and other complex environments. Despite such successes, deep RL algorithms are sample inefficient and sometimes unstable. Furthermore, they usually perform sub-optimally when dealing with sparse reward and partially observable environments. One further limitation of deep RL is when rapid adaptation to changing tasks (dynamic goals) is required. Established methods only work well in fixed task environments. In an attempt to solve this problem, deep meta-reinforcement learning (meta-RL) methods \cite{finn2017model, rothfuss2018promp, zintgraf2019fast, duan2016rl, wang1611learning} were specifically devised. However, these methods are largely evaluated on dense reward, fully observable MDP environments, and perform sub-optimally in sparse reward, partially observable environments.

One key aspect in achieving fast adaptation in dynamic partially observable environments is the presence of appropriate learning structures and memory units that fits the specific class of learning problems. Therefore, standard model-free RL algorithms do not perform well in dynamic environments because they are tabula-rasa systems. They hold no knowledge in their architectures to allow a fast and targeted learning when a change in the environment occurs. Upon a task change, these algorithms will try to randomly explore the action space to relearn from scratch a different, new policy. On the other hand, model-based RL, holds knowledge of the structure of the environment, which in turn allows for rapid adaptation to changes in the environment, but such a knowledge needs to be built manually into the system. 

In this paper, we investigate the use of neuroevolution to autonomously evolve inborn knowledge \cite{soltoggio2018born} in the form of neural structures and plasticity rules with a specific focus on dynamic POMDPs that have posed challenges to current RL approaches. The neuroevolutionary approach that we propose is designed to solve rapid adaptation to changing tasks \cite{soltoggio2018born} in complex high dimensional partially observable environments. The idea is to test the ability of evolution to build an unconstrained neuromodulated network architecture with problem-specific learning skills that can exploit the latent space provided by an autoencoder. Thus, in the proposed system, an autoencoder serves as a feature extractor that produces low dimensional latent features from high dimensional environment observations. A neuromodulated network \cite{soltoggio2008evolutionary} receives the low dimensional latent features as input and produces the output of the system, effectively acting as high level controller. Evolved neuromodulated networks have shown computational advantages in various dynamic task scenarios \cite{soltoggio2008evolutionary, soltoggio2018born}.

The proposed approach is similar to that proposed in \cite{alvernaz2017autoencoder}. One key novelty is that our approach seeks to evolve selective plasticity with the use of modulatory neurons, and therefore, to evolve problem-specific neuromodulated adaptive systems. The relationships among image-pixel inputs and control actions in POMDPs is highly nonlinear and history dependent, therefore, an open question is whether neuroevolution can exploit latent features to evolve learning systems with inborn knowledge. Thus, we test the hypothesis that a neuromodulated evolved network can discover neural structures and their related plasticity rules to encode required memory and fast adaptation mechanisms to compete with current deep meta-RL approaches.

We call the proposed system a Plastic Evolved Neuromodulated Network with Autoencoder (PENN-A), denoting the combination of the two neural components. We evaluate our proposed method in a POMDP environment where we show better performance in comparison to some non-evolutionary deep meta-reinforcement learning methods. Also, we evaluated the proposed method in the Malmo Minecraft environment to test its general applicability.

Two interesting findings from our experiments are that   (i) the networks acquire through evolution the ability to recognise reward cues (i.e. environment cues that are associated with survival even when reward signals are not given) and (ii) the networks can evolve \emph{location} neurons that help solving the problem by detecting, and becoming active at, specific location of the partially observable MDP. The evolved network topology allows for richer dynamics in comparison to fixed architectures such as hand-designed feed-forward or recurrent networks.

The next section reviews the related work. Following that, a formal task definition is presented. Next is the description of the proposed method employed in this work, followed by the evaluation of results. The PENN-A source code is made available at: \url{https://github.com/dlpbc/penn-a}.

\section{Related Work}
In reinforcement learning (RL) literature, meta-RL methods seek to develop agents that adapt to changing tasks in an environment or a set of related environments. Meta-RL \cite{schmidhuber1996simple, schweighofer2003meta} is based on the general idea of meta-learning \cite{bengio1992optimization, thrun1998learning, hochreiter2001learning} applied to the RL domain.

Recently, deep meta-RL has been used to tackle the problem of rapid adaptation in dynamic environments. Methods such as \cite{finn2017model, duan2016rl, wang1611learning, zintgraf2019fast, mishra2018a, rothfuss2018promp, rakelly2019efficient} use deep RL methods to train a meta-learner agent that adapts to changing tasks. These methods are mostly evaluated in dense reward, fully observable MDP environments. Furthermore, most methods are either memory based \cite{duan2016rl, wang1611learning, mishra2018a} or optimization based \cite{finn2017model, zintgraf2019fast}. Optimization based methods seek to find an optimal initial set of parameters (e.g. for an agent network) across tasks, which can be fine-tuned with a few gradient steps for each specific task presented to it. Therefore, a small amount of re-training is required to enable adaptation to every change in task. Memory based methods (implemented using a recurrent network or temporal convolution attention network) do not necessarily require fine tuning after initial training to enable adaptation. This is because memory-based agents learn to build a memory of past sequence of tasks and interactions, thus enabling them to identify change in task and adapt accordingly.

In the past, neuroevolution methods have been employed to solve RL tasks \cite{stanley2002evolving, mchale2004gasnets}, including adapting to changing tasks \cite{soltoggio2008evolutionary, blynel2002levels} in partially observable environments. These methods were evaluated in environments with high level feature observations. Recently, several approaches have been introduced that combine deep neural networks and neuroevolution to tackle high dimensional deep RL tasks \cite{alvernaz2017autoencoder, poulsen2017dlne, ha2018recurrent, salimans2017evolution, such2017deep}. These approaches can be divided into two major categories. The first category uses neuroevolution to optimize the entire deep network end to end \cite{salimans2017evolution, such2017deep, risi2019deep, risi2019improving}. The second category splits the network into parts (for example, a body and controller) where some part(s) (e.g. body) are optimized using gradient based methods and other part(s) (e.g. controller) are evolved using neuroevolution methods \cite{alvernaz2017autoencoder, poulsen2017dlne, ha2018recurrent}. Current deep neuroevolution methods are usually evaluated in fully observable MDP environments, where the task is fixed. Furthermore, after the training phase is completed, the weights of a trained network are fixed (the same is true for standard deep RL). The recent attention to neuroevolution for deep RL aims to present such approaches as a competitive alternative to standard gradient based deep RL methods for fixed task problems.

In the past, neural network based agents employing Hebbian-based local synaptic plasticity have been used to achieve behavioural adaptation with changing tasks \cite{floreano1996evolution, blynel2002levels, soltoggio2008evolutionary}. Such methods use a neuroevolution algorithm to optimize the parameters of the network when producing a new generation of agents. As an agent interacts with an environment during its lifetime in training or testing, the weights are adjusted in an online fashion (via a local plasticity rule), enabling adaptation to changing tasks. In \cite{floreano1996evolution, blynel2002levels} this technique was employed, and further extended to include a mechanism of gating plasticity via neuromodulation in \cite{soltoggio2008evolutionary}. These methods were evaluated in environments with low dimensional observations (with high level features) and not compared with deep (meta-)RL algorithms. 

\section{Task Definition}
\label{sec:task-definition}
A POMDP environment $E$, defined by a sextuple ($\mathcal{S}$, $\mathcal{A}$, $\mathcal{P}$, $\mathcal{R}$, $\mathcal{O}$, $\Omega$) is employed in this work. $\mathcal{S}$ defines the state set, $\mathcal{A}$ the action set, $\mathcal{P}: \mathcal{S} \times \mathcal{A} \times \mathcal{S} \to [0, 1]$ the environment dynamics, $\mathcal{R}: \mathcal{S} \times \mathcal{A} \to \mathbb{R}$ the reward function, $\mathcal{O}$ the observation set, and $\Omega$ the function that maps observations to states.

The environment $E$ contains a number of related tasks. A task $\mathcal{T}_{i}$ is sampled from a distribution of tasks $\mathcal{T}$. The task distribution $\mathcal{T}$ can either be discrete or continuous. A sampled task is an instance of the partially observable environment $E$. The configuration of the environment (for example, the goal or reward function) varies across each task instance. An optimal agent is required to adapt its behaviour to task changes in the environment (and maximize accumulated reward), only from few interactions in the environment. When presented with a task $\mathcal{T}_i$, an optimal agent should initially explore, and subsequently exploit when the task is understood. When the task is changed (a new task $\mathcal{T}_j$ sampled from $\mathcal{T}$), the agent needs to re-explore the environment in few-shots, and then to start exploiting again when the new task has been understood.

\begin{figure}
    \centering
    \includegraphics[scale=0.6]{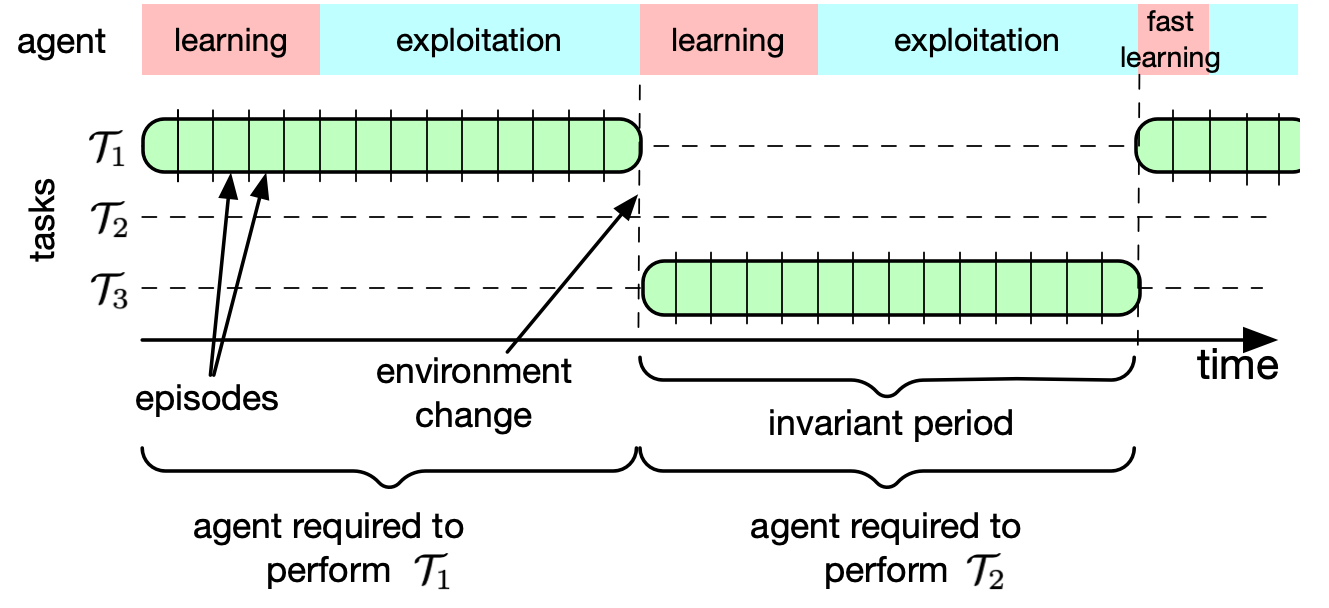}
    \caption{Illustration of a dynamic environment and required behavior of a learning agent. An agent is required to learn to perform optimally and then exploit the learned policy until a change in the environment occurs, at which point the agent needs to learn again before exploiting.}
    \label{fig:fig-task-change}
\end{figure}

In each task, an episode is defined as the trajectory $\tau$ of an agent interactions in the environment, terminating at a terminal state. A trial consist of two or more tasks sampled from $\mathcal{T}$. The total number of episodes in a trial is kept fixed. A trial starts with an initial task $\mathcal{T}_i$ that runs for a number of episodes, and then the task is changed to other tasks (one after another) at different points within the trial (see Figure \ref{fig:fig-task-change}). The points at which a task change occurs are stochastically generated, and the task is changed before the start of the next episode. For example, when the number of tasks is set as $2$ (i.e. $\mathcal{T}_i$ and $\mathcal{T}_j$), the trial starts with task $\mathcal{T}_i$ which runs for a number of episodes, and it is replaced by task $\mathcal{T}_j$ for the remaining episodes in the trial. An agent is iteratively trained, with each iteration consisting of a fixed number of trials. The subsections below describes two environments where the proposed system is evaluated.

\begin{figure*}
  \includegraphics[width=\textwidth]{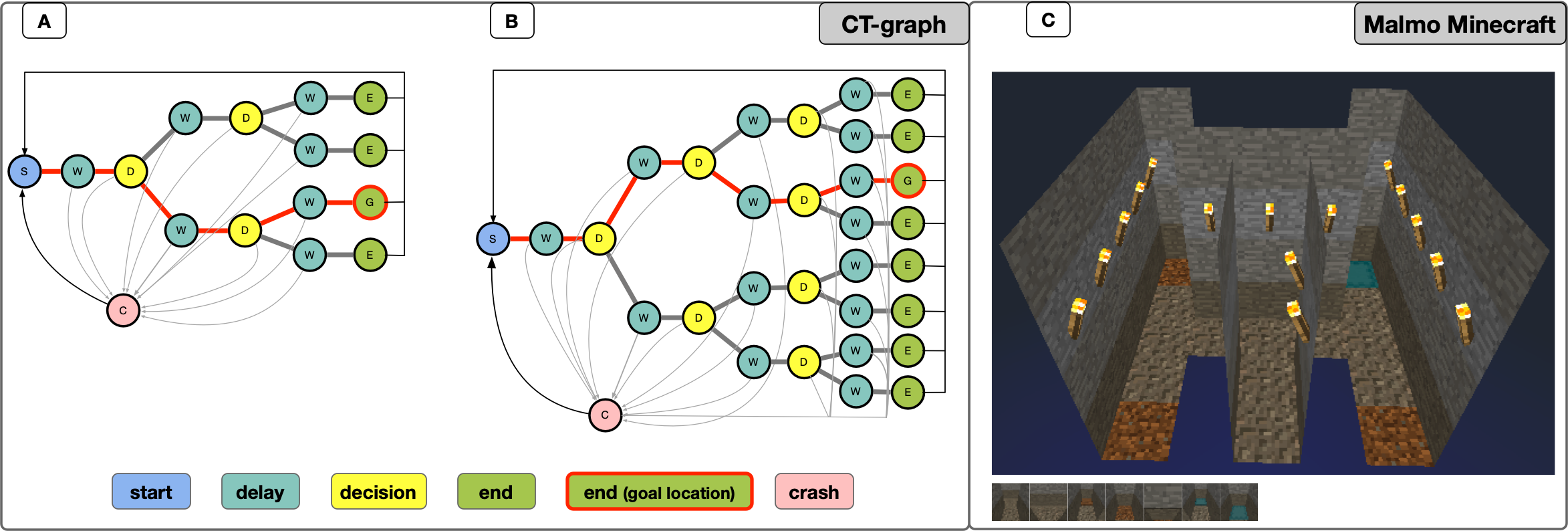}
  \caption{Environments (note, during execution, goal location is dynamic across episodes). (A) CT-graph instance, $b=2$ and $d=2$. (B) CT-graph instance, $b=2$ and $d=3$. (C) Malmo Minecraft instance (a double T-Maze), bird's eye view on top, with some sample observations at the bottom. The maze-end with the teal colour is the goal location.}
  \label{fig:fig-ctgraph-minecraft-env}
\end{figure*}

\subsection{The Configurable Tree Graph Environment}
The configurable tree graph (CT-graph) environment is a graph abstraction of a decision making process. The complexity of the environment is specified via configuration parameters; branching factor $b$ and depth $d$, controlling the width and height of the graph. Additionally, it can be configured to be fully or partially observable. It contains the following types of state; start, wait, decision, end (leaf node of graph) and crash. Each observation $o \in \mathcal{O}$ is a $12x12$ grey-scale image. The total number of end states grows exponentially as the depth $d$ of the graph increases (see Figure \ref{fig:fig-ctgraph-minecraft-env}A and B).

In the experiments in this study, partial observability is configured by mapping all wait states to the same observation, and all decision states to the same observation. Also, $b$ is set to 2. Therefore, each decision state has two choices, splitting into two sub-graphs. The discrete action space is defined as; \textit{choice 1, choice 2, wait action},
thus discrete. The \textit{wait action} is the correct action in a wait state. In a decision state, \textit{choice 1} or \textit{choice 2} is the correct subset from which to select. All incorrect actions lead to the crash state and episode termination.

An agent starts an episode in the start state, and the episode is completed when the agent traverses the graph to an end state or takes a wrong action in a state. Once an agent transitions from one state to the next, it cannot go back. In a task instance, one of the end states is set as the goal location. An agent receives a positive reward when it traverses to the goal location, and reward of 0 at other non-goal states. The agent may receive a negative reward in a crash state.

\subsection{Malmo Minecraft Environment}
Malmo \cite{johnson2016malmo} is an AI research platform built on top of Minecraft video game. The platform is configurable, and it enables the construction of various worlds in which AI agents can be evaluated. In this work, a double T-maze was constructed, with discrete action space \textit{left turn, right turn and forward action}. A task is defined based on the maze ends, requiring the agent to navigate to a specific maze end (goal location). The maze end that is set as the goal location varies across tasks. The agent only receives a positive reward when they navigate to the maze end that is the goal location. It receives reward of 0 in every other time step. If the agent runs into a wall, the episode is terminated and it receives a negative reward. The agent receives a visual observation of its current view at each time step (hence it does not fully observe the entire environment). Each observation is a $32 x 32$ RGB image based on a first-person view of the agent at each time step.

\section{Methods}
\label{sec:methods}
\begin{figure}
  \includegraphics[scale=1.2]{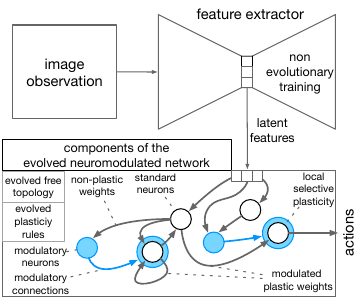}
  \caption{System overview, showcasing the feature extractor and controller components. In the controller, white and blue nodes are standard and modulatory neurons respectively. Modulatory connections facilitates selective plasticity in the network.}
  \label{fig:fig-system-overview}
\end{figure}

We seek to develop an agent that is capable of continual adaptation through its life time (across episodes) - exploring, exploiting, re-exploring when the task changes and exploiting again. The system (specifically the controller or decision maker) is evolved to acquire knowledge about both the invariant and variant aspects of an environment (e.g. changing tasks).

The agent is modelled using two neural components with separate parameters and objectives; a deep network $F_\theta$ (used as a feature extractor and parameterized by $\theta$) and a neuromodulated network $G_\phi$ (serving as a controller and parameterized by $\phi$). Both components make up the overall system model $\mathcal{M_{\theta, \phi}}$. See Figure \ref{fig:fig-system-overview} for a general system overview. The presented architectural style is similar to a standard deep RL setup. However, it differs on two fronts; (i) the controller is a neuromodulated network (described in Section \ref{sec:controller}) rather than a standard neural network, (ii) the training setup combines gradient based optimization method \cite{werbos1982applications, rumelhart1988learning}), gradient free optimization method (neuroevolution \cite{yao1999evolving, stanley2019designing}), and Hebbian-based synaptic plasticity to train the system. Using this setup, each neural component therefore contains its own objective function. An autoencoder network was employed as the feature extractor, thus enabling the use of Mean Squared Error (MSE) or Binary Cross Entropy (BCE) objective function:
\begin{displaymath}
    \argmin_{\theta} \frac{1}{n} \sum_{i=1}^{n} (F_{\theta}(o_i) - o_i)^2
\end{displaymath}
\begin{displaymath}
    \argmin_{\theta} - \frac{1}{n} \sum_{i=1}^{n} o_i \cdot \log({F_{\theta}(o_i)}) + (1 - o_i) \cdot \log{(1 - F_{\theta}(o_i))}
\end{displaymath}
where $n$ is the number of training observations and $F_{\theta}(o_i)$ is the output of the autoencoder for observation $i$ (reconstructed observation). Each agent in the population uses the same feature extractor. The fitness function of the evolutionary algorithm is given by:

\begin{displaymath}
    \argmax_{\phi} \sum_{\mathcal{T}_i \sim \mathcal{T}} \sum_{ep=1}^{z} R(\tau_{ep})
    \label{eq:fitness}
\end{displaymath}
$\mathcal{T}_i$ represents a task sampled from the task distribution $\mathcal{T}$, and a single trial consist of two tasks as defined in Section \ref{sec:task-definition}. Also, z is the number of episodes in which a task is kept fixed within a trial. It is stochastically generated and may differ between tasks in a trial within an interval. $R(\tau_{ep})$ is the accumulated reward of a trajectory of an episode $ep$, defined as:

\begin{equation}
    R(\tau_{ep}) = \sum_{t=0}^{k} \mathcal{R}(s_t, G_\phi(F_\theta^{enc}(o_t)))
\end{equation}
where $\mathcal{R}(s, a)$ is the reward function that takes state and action as arguments and produces a scalar reward value. $F_{\theta}^{enc}$ is the same autoencoder feature extractor network earlier described, but denoting that we only want the output from the encoder (the latent features). Also, $t$ represents discrete time steps and $k$ is the length of the trajectory of an episode.

\subsection{Feature Extractor}
\label{sec:feature_extractor}
This neural component of the system is tasked with learning a good latent representation of the observations from the environment, which can be fed to the controller as input. In the CT-graph experiments, a fully connected autoencoder was employed (two layers encoder and decoder respectively). In the Malmo Minecraft experiments, a convolutional autoencoder was employed (four layers encoder and decoder respectively).

\subsection{Control Network (Decision Maker)}
\label{sec:controller}
This neural component takes the latent features of the feature extractor as its input, and produces an output which serves as the final output of the system (the action or behaviour of the system). It is a neuromodulated network (see Section \ref{sec:neuromodulated-network-dynamics}), that reproduces the model introduced in \cite{soltoggio2008evolutionary}. The network can evolve two neuron types - a standard and a modulatory neuron. The output neuron(s) always belong to the standard neuron type.

The control network is parameterized by $\phi$. Unlike $\theta$ (which represents only the weights of the feature extractor network), $\phi$ consists of the weights, architecture and the co-efficients of Hebbian-based plasticity rule (described in \ref{sec:neuromodulated-hebbian-plasticity}) of the network, and it is evolved. Therefore, evolution is tasked with finding the architecture and plasticity rules, including selective plasticity enabled by modulatory neurons to target neurons. The large search space that is granted to evolution allows for rich dynamics that include memory in the form of both recurrent connections and temporary values of rapidly changing modulated weights.

The agent is never fed the reward signal explicitly. The reward signal is only used by the evolutionary process for the fitness evaluation, which in turn drives the selection process. Therefore, the network is tasked to learn the discovery of reward cues implicitly from the visual observations in the environment.

\subsubsection{Neuromodulated Network Dynamics}
\label{sec:neuromodulated-network-dynamics}
Though processing is distributed across neurons, a standard neural network usually contains one type of neuron - where the dynamics of each neuron is homogeneous across the network. In a neuromodulated network, there can be two types of neurons, each type having different dynamics - thus heterogeneous. The two types of neurons are standard neurons and modulatory neurons \cite{soltoggio2008evolutionary}. The standard neurons have the same dynamics as the ones in standard neural network. The modulatory neurons are used to dynamically regulate plasticity in the network.

Each neuron $i$ has one standard and one modulatory activation value that represent the weighted amount of standard and modulatory activity they receive from other neurons (see Equations \ref{eq:eq-activation-std} and \ref{eq:eq-activation-mod}). $a_{std,i}$ is the output signal of neuron $i$ that is propagated to other neurons in its outgoing connections (this is true for both standard and modulatory neurons). $a_{mod,i}$ is used internally by the neuron itself to regulate the Hebbian-based plasticity of the incoming connections from other standard neurons, as described in Section \ref{sec:neuromodulated-hebbian-plasticity}. The framework allows for selective plasticity in the network, as parts of the network may become plastic or not plastic depending on the change of the modulatory activation signals over time. In turn, the final action of the network is affected in the current and future time steps - thus enabling adaptation.

\begin{equation}
    a_{\mathrm{std},i} = \tanh{ \frac{\sum_{j \in \mathrm{std}} w_{ji}a_{\mathrm{std},j}}{2} } 
    \label{eq:eq-activation-std}
\end{equation}
\begin{equation}
    a_{\mathrm{mod},i} = \tanh{ \frac{\sum_{j \in \mathrm{mod}} w_{ji}a_{\mathrm{std},j}}{2} }
    \label{eq:eq-activation-mod}
\end{equation}

\subsubsection{Neuromodulated Hebbian Plasticity}
\label{sec:neuromodulated-hebbian-plasticity}
The Hebbian synaptic plasticity of the control network is governed by the Equations \ref{eq:eq-weight-update}, \ref{eq:eq-modulated-weight-delta} and \ref{eq:eq-weight-delta}. $A, B, C, D, \alpha$ are the coefficients of the plasticity rule. The update of a weight is dependent pre-synaptic and post-synaptic standard activations, the plasticity co-efficients, and the post-synaptic modulatory activation. This is true for all weights in the neuromodulated network.

\begin{equation}
    w_{ij} = w_{ij} + \Delta w_{ij}
    \label{eq:eq-weight-update}
\end{equation}

\begin{equation}
    \Delta w_{ij} = a_{\mathrm{mod}, j} \cdot \delta w_{ij}
    \label{eq:eq-modulated-weight-delta}
\end{equation}

\begin{equation}
    \delta w_{ij} = \alpha \cdot (A \cdot a_{\mathrm{std}, i} \cdot a_{\mathrm{std}, j} + B \cdot a_{\mathrm{std}, i} + C \cdot a_{\mathrm{std}, j} + D)
    \label{eq:eq-weight-delta}
\end{equation}

\section{Results and Analysis}
\begin{figure}
  \includegraphics[scale=0.45]{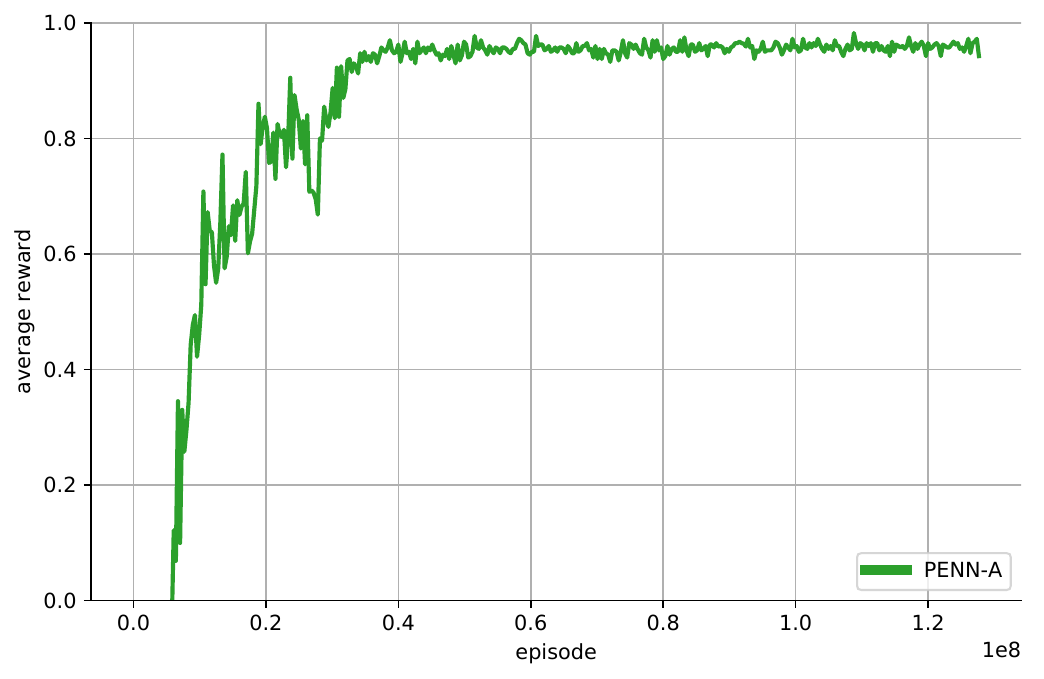}
  \caption{Malmo Minecraft result.}
  \label{fig:minecraft-double-tmaze-results}
\end{figure}

Figures \ref{fig:ctgraph-depth2-results} and \ref{fig:ctgraph-depth3-results} show the results of the experiments in the CT-graph environment. Figure \ref{fig:minecraft-double-tmaze-results} shows the results of the experiment in the Malmo Minecraft environment. In addition, we present results obtained in the Malmo Minecraft environment (Figure \ref{fig:minecraft-double-tmaze-results}), evaluating the general applicability of PENN-A.

\subsection{Performance in CT-graph Environments}
The proposed method (PENN-A) was evaluated on depth 2 and 3 CT-graph environments, with branching factor of 2. The controller was evolved for 200 generations, with population of 600 and 800 for depth 2 and 3 experiments respectively. Tournament selection with segment size of 5 was employed. Each controller was evaluated for 4 trials, with 100 episodes and 2 tasks per trial. The initial task is changed between episodes 35 and 65, determined stochastically for each trial. The depth 2 CT-graph experiment was employed as a baseline, and we compared PENN-A against some recent deep meta-RL methods (each with its own experimental setup). The depth 3 CT-graph experiment was employed to evaluate the PENN-A in a more complex configuration of the environment.

\begin{figure}
  \includegraphics[scale=0.45]{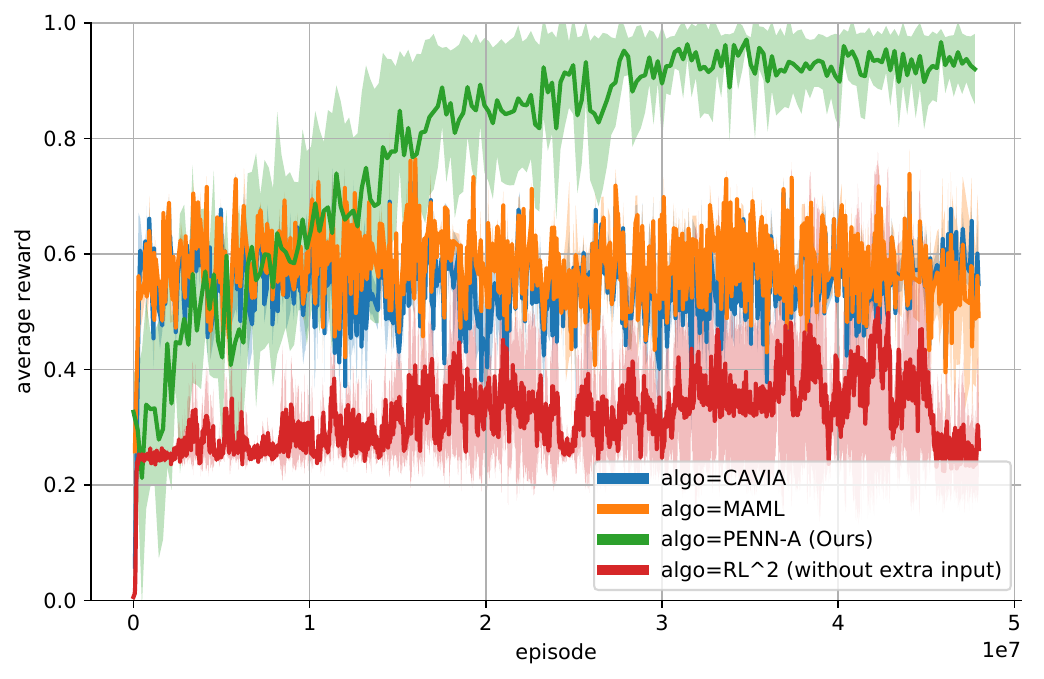}
  \caption{Results for a CT-graph with depth 2. PENN-A is compared against non-evolutionary meta-RL methods.}
  \label{fig:ctgraph-depth2-results}
\end{figure}

\begin{figure}
  \includegraphics[scale=0.45]{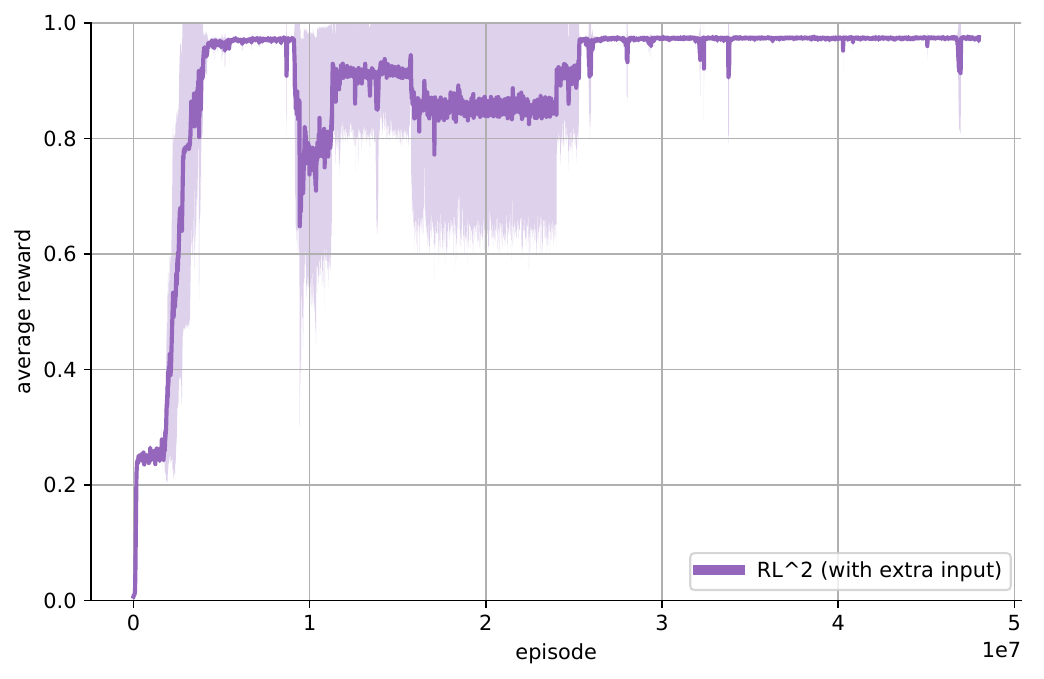}
  \caption{$\text{RL}^2$ in the CT-graph with depth 2. The method is run with extra input to the network (reward, done state, and previous time step action concatenated with current observation to form input).}
  \label{fig:ctgraph-depth2-results-rl2-with-extra-input}
\end{figure}

In order to ensure compatibility in the result presented across all methods, the number of evaluations (horizontal axis) were scaled to the approximate number of episodes equivalent. Additionally, the vertical axis is the average accumulated reward across all trials and episodes. In the depth 2 CT-graph result (Figure \ref{fig:ctgraph-depth2-results}), we see that PENN-A performs optimally when compared to deep meta-RL methods; optimization-based (MAML \cite{finn2017model} and CAVIA \cite{zintgraf2019fast}) and memory-based ($\text{RL}^2$ \cite{duan2016rl} without extra input). Only the observations were fed as input to the neural network for all methods including PENN-A. We hypothesize the deep meta-RL methods perform sub-optimally due to the partial observability of the environment. When extra input (the reward, previous time step action and done state) are concatenated to the observation and fed to the $\text{RL}^2$ method (which is vanilla setup), then it is able to perform optimally (see Figure \ref{fig:ctgraph-depth2-results-rl2-with-extra-input}). We hypothesize that $\text{RL}^2$ exploits the actions fed as input to the network, ignoring the observations and other parts of the input. This reduces the problem complexity in comparison to conditions where only the observations are fed as input.

\begin{figure}
  \includegraphics[scale=0.45]{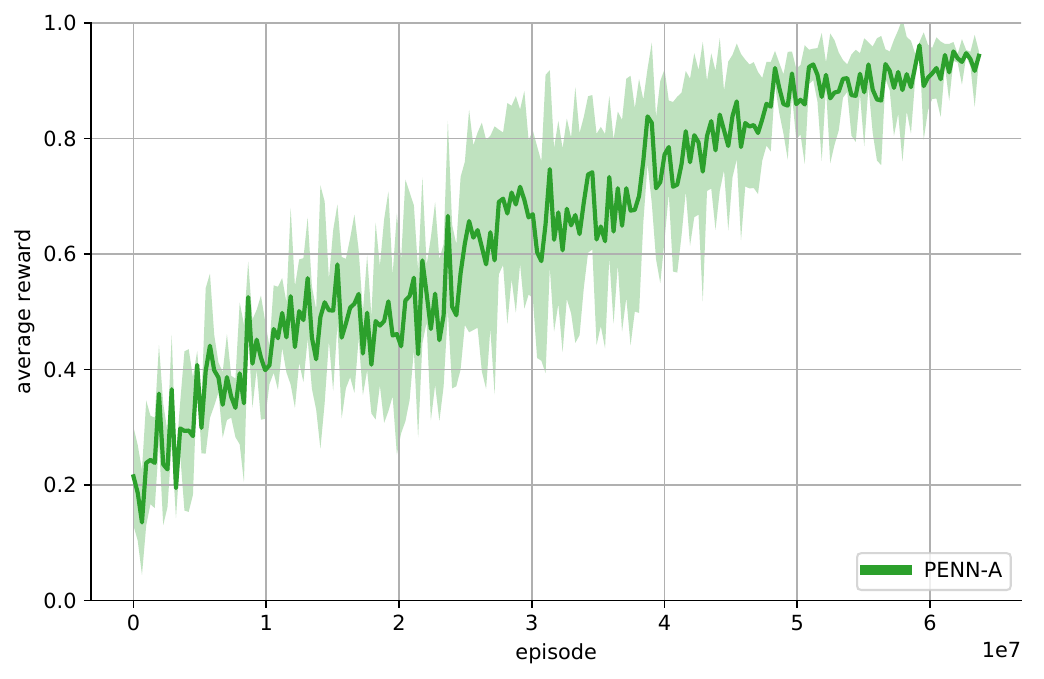}
  \caption{PENN-A performance in a CT-graph with depth 3.}
  \label{fig:ctgraph-depth3-results}
\end{figure}

Figure \ref{fig:ctgraph-depth3-results} presents result for a depth 3 CT-graph. We present result for only PENN-A in depth 3 CT-graph (a more difficult problem than depth 2 CT-graph) since the other methods performed sub-optimally in depth 2 CT-graph. We again observe PENN-A performing optimally in the more difficult CT-graph setting.

\subsubsection{Network Analysis}
\begin{figure}
  \includegraphics[scale=0.45]{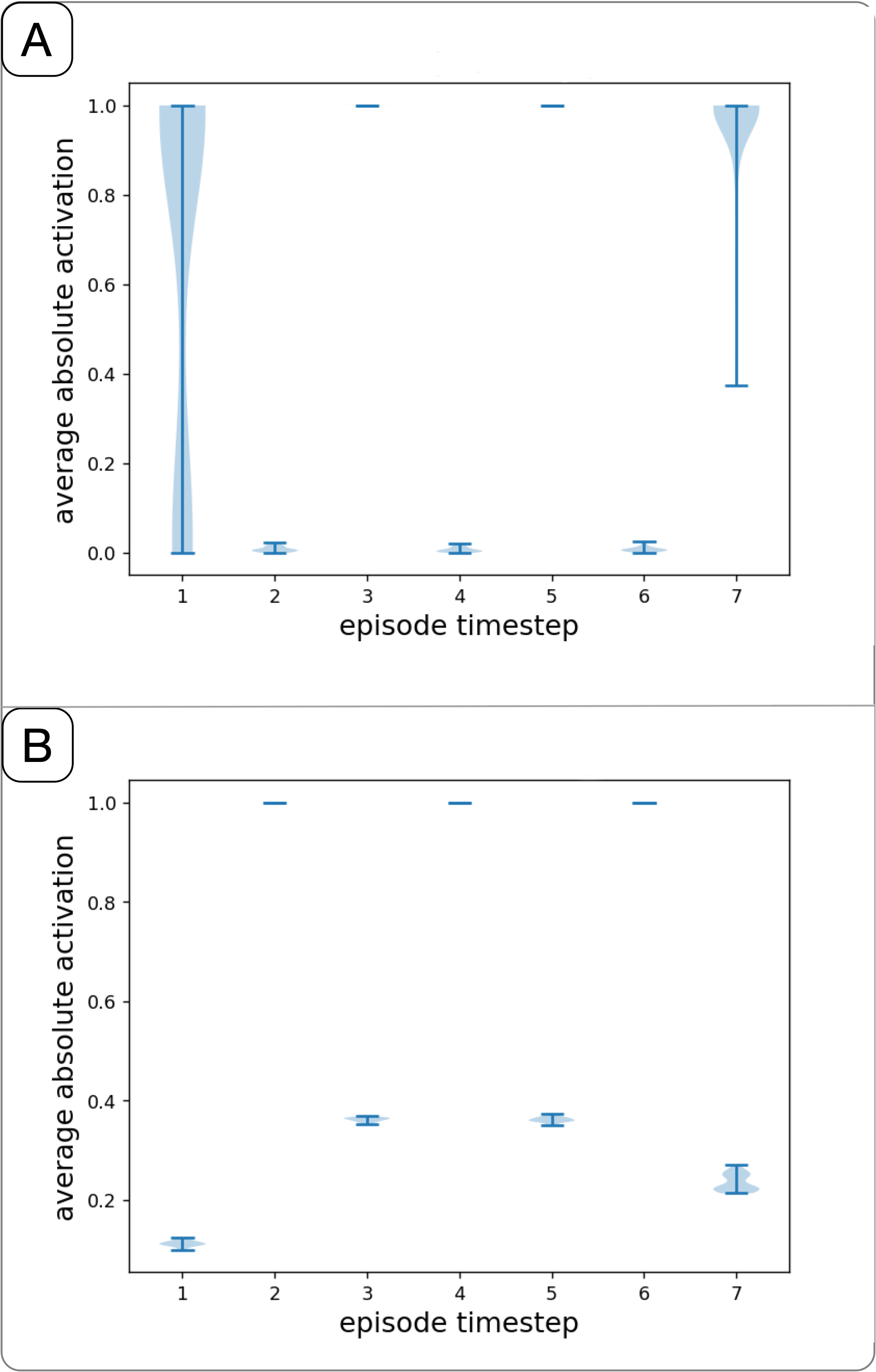}
  \caption{Absolute activation values distribution (across trials and episodes) per time step of a sample evolved controller. (A) This neuron is active specifically at decision states (steps 3 and 5), while it remains low at wait states. (B) This neuron clearly identifies wait states (steps 2, 4 and 6) and remains inactive otherwise.}
  \label{fig:fig-average-absolute-neurons-activation}
\end{figure}

\begin{figure*}
  \includegraphics[width=\textwidth]{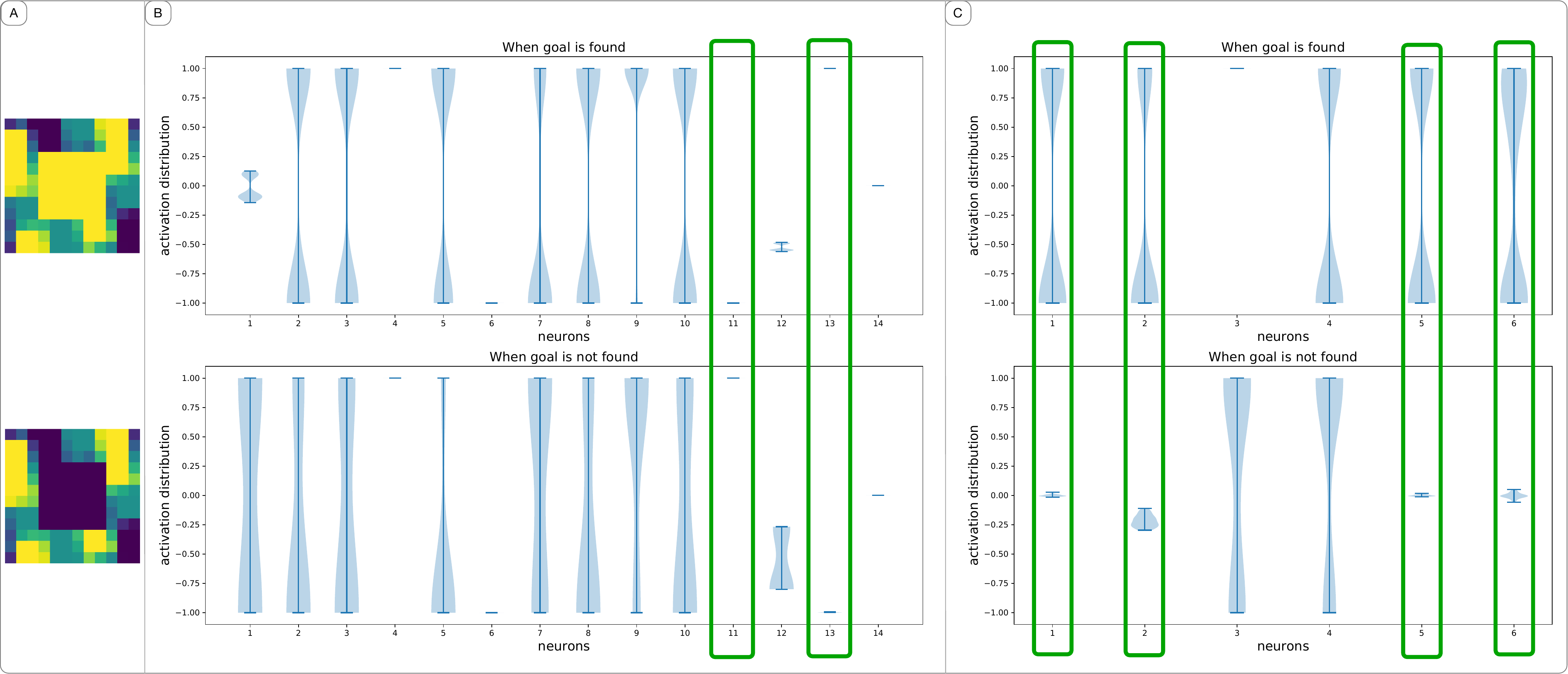}
  \caption{Distributions of the activation values for each neuron in a sample network when the goal location (reward) is found and vice versa. The neurons highlighted in green bounding box react differently to the presence or absence of reward cues from observation. (A) heat maps of grey-scale CT-graph observations. The top image is the observation presented when the goal location is found, with a bright square reward cue. The bottom image is the observation when the goal location is not found, with the reward cue absent. (B) Neurons 11 and 13 show complementary firing patterns based on reward cues. (C) Neurons 1, 2, 5 and 6 are active when a reward cue is observed, and have little or no activity when the reward cue is not observed.}
  \label{fig:fig-reward-cues-neurons-activation}
\end{figure*}

To better understand the evolved solution and how the network implements policies, we analyzed the best performing networks after evolution in a depth 2 CT-graph environment. While different evolutionary runs produced highly different networks, we observed interesting patterns in the neural activations. For one network of 11 neurons (including the output neuron), the absolute activation value distribution (across trials and episodes per time step) is plotted for each neuron in Figure \ref{fig:fig-average-absolute-neurons-activation}. We see that the absolute activation distribution of some neurons are high at specific time steps, i.e., at specific points within the graph environment (see Figure \ref{fig:fig-average-absolute-neurons-activation}A and B) -  and therefore function as \emph{location neurons}. Such kind of location neurons had been previously discovered in an evolutionary setting in \cite{floreano2010evolution}. In the current experiments, it is worth noting that location neurons are designed by evolution to exploit latent features and possibly help action-selection in a high-dimensional dynamic POMDP. In particular, the neuron in Figure \ref{fig:fig-average-absolute-neurons-activation}A is active at decision states, while the neuron in Figure \ref{fig:fig-average-absolute-neurons-activation}B is active at wait states.

One aspect of our experimental setting is that the reward signal is not fed to the network, but the environment provides reward cues embedded in the observations as it is shown in Figure \ref{fig:fig-reward-cues-neurons-activation}A where a bright square represents a reward. The actual reward value is only accumulated in the fitness function, and is therefore not explicitly visible to the network. The surprising results that networks evolved to explore the environment and find the reward even if no reward signal was given suggests that the reward cue was recognised. In fact, in the example shown in Figure \ref{fig:fig-reward-cues-neurons-activation}B, some neurons fire positively when a reward cue is observed and negatively when not observed or vice versa. Other neurons fire when a reward cue is observed and have little or no firing when not observed (see Figure \ref{fig:fig-reward-cues-neurons-activation}C). Not all evolved networks appeared to have \emph{reward neurons}. Nevertheless, the examples that evolved such reward cues detectors demonstrate that evolution is able to incorporate invariant knowledge of the environment to optimize the policy, in this case, reward seeking behaviour and fast adaptation speed to changing task.

\subsection{Performance in Malmo Minecraft}
To further assess the validity of our method, it is important to use a different benchmark environment with a larger input and RGB observations that offered a different feature space, hence the Malmo Minecraft environment. The controller was evolved with population size of 800, in 400 generations. The same selection strategy as used in the CT-graph was employed. Each controller was evaluated for 8 trials, with 50 episodes and 3 tasks per trial. The task is changed at two stochastically generated points within the trial. The result is presented in Figure \ref{fig:minecraft-double-tmaze-results}, keeping the same axes format as with the results presented for the CT-graph environment. Again, the proposed method was able to perform optimally with a high average reward score, demonstrating its capability to scale to other high dimensional, less abstract environments.

\section{Conclusion}
\label{sec:conclusion}
This paper introduced an evolutionary design method for fast adaptation in POMDP environments. The system combines a feature extractor network and an evolved neuromodulated network with the aim of acquiring specific inborn knowledge and structure via evolution. While the suitability of evolved neuromodulated networks to solve environments with changing task was known \cite{soltoggio2008evolutionary, soltoggio2018born}, we demonstrated that such advantages are scalable to high dimensional input spaces, and can be used in combination with an autoenconder. The results showed performance that compare or surpass some deep meta-RL algorithms. Interestingly, the evolved networks were capable of learning to recognise implicit reward cues, and therefore could explore the environment in search for the goal location without an explicit reward signal. This ability that was acquired by the networks through evolution is an example of inborn knowledge that allow networks to be born with the knowledge of what are reward cues. Subsequently, this information can be used to direct fast adaptation when the optimal policy changes (e.g. the task change). The networks also evolved location neurons to help the deployment of a policy by distinguishing different states in the underlying MDP. We speculate that this approach might be promising when a combination of inborn knowledge and online learning are required to perform optimally in rapidly changing environments.

\appendix
\section{Experimental Settings}
\label{appendix:other-experimental-configurations}
The PENN-A source code containing the experimental setup is made available at: \url{https://github.com/dlpbc/penn-a}.

\subsection{Feature Extractor}
\label{appendix:feature-extractor}
Mean Squared Error (MSE) was employed as the loss function across all CT-graph experiments, with a vanilla SGD (learning rate of $0.001$) optimizer. Likewise, Binary Cross Entropy (BCE) was employed as the loss function in the Malmo Minecraft experiments, using RMSprop (learning rate of $0.0005$) optimizer. The network architectures for the CT-graph and Malmo Minecraft experiments are presented in Table \ref{tab:tbl-fe-net-archi-ctgraph} and \ref{tab:tbl-fe-net-archi-minecraft}. The double horizontal line in both tables highlight the split between the encoder and decoder (i.e. specifications on top of the double line are for the encoder and likewise bottom for the decoder) of each autoencoder. In the CT-graph experiments where a Fully Connected (FC) autoencoder was employed, each input observation is flattened into a vector before feeding it to the network.

\begin{table}
  \caption{Network architecture for CT-graph experiments}
  \label{tab:tbl-fe-net-archi-ctgraph}
  \begin{tabular}{lcc}
    \toprule
    Layer & Activation & Units\\
    \midrule
    Input & N/A & 144\\
    \hline
    FC & ReLU & 64\\
    FC & ReLU & 16\\
    \hline \hline
    FC & ReLU & 64\\
    FC & ReLU & 144\\
  \bottomrule
\end{tabular}
\end{table}

\begin{table}
  \caption{Network architecture for Malmo Minecraft experiments}
  \label{tab:tbl-fe-net-archi-minecraft}
  \begin{tabular}{lcccc}
    \toprule
    Layer & Activation & Kernel & Stride & Channels\\
    \midrule
    Input & N/A & N/A & N/A & N/A\\
    \hline
    Conv2D & ReLU & 3x3 & 2 & 16\\
    Conv2D & ReLU & 3x3 & 2 & 32\\
    Conv2D & ReLU & 3x3 & 2 & 32\\
    Conv2D & ReLU & 3x3 & 2 & 8\\
    \hline \hline
    ConvTranspose2D & ReLU & 3x3 & 2 & 32\\
    ConvTranspose2D & ReLU & 3x3 & 2 & 32\\
    ConvTranspose2D & ReLU & 3x3 & 2 & 16\\
    ConvTranspose2D & Sigmoid & 4x4 & 2 & 3\\
  \bottomrule
\end{tabular}
\end{table}

\subsection{Control Network}
\label{appendix:controller-network}
Excluding population size and number of generations, the evolutionary parameters from \cite{soltoggio2008evolutionary} were followed.

The latent features from the feature extractor network were scaled between $0$ and $1$. To further restrict the latent features, a transformation operation was applied to the scaled latent features $v$ before it was fed to the control network as shown below.

\begin{displaymath}
    w = \left\{
    \begin{array}{cc}
         1 & if \sigma^{\prime}(v) > 1  \\
         0 & if \sigma^{\prime}(v) < 0 \\
         \sigma^{\prime}(v) & otherwise
    \end{array} \right.
\end{displaymath}
\begin{displaymath}
    \sigma^{\prime}(v) = \log \frac{v}{1-v}
\end{displaymath}
where $\sigma^{\prime}(v)$ is an inverse sigmoid operation on $v$, and $w$ is the transformed feature space. The scaling and transformation operations were performed independently of the feature extractor optimization (i.e. the operations were applied on copies of the latent features), and were applied across all experiments.

In this work, both evaluation environments were designed to work with discrete action space (3 actions each). Therefore, a single output neuron was employed across all experiments. The $\tanh$ activation value of the neuron was discretized to produce the actions of an agent. An activation value within the interval [-1.0, -0.33) mapped to one action, the interval [-0.33, +0.33] mapped to another action, and the interval (+0.33, 1.0] mapped to the last action.

\begin{acks}
This material is based upon work supported by the United States Air Force Research Laboratory (AFRL) and Defense Advanced Research Projects Agency (DARPA) under Contract No. FA8750-18-C-0103. Any opinions, findings and conclusions or recommendations expressed in this material are those of the author(s) and do not necessarily reflect the views of the United States Air Force Research Laboratory (AFRL) and Defense Advanced Research Projects Agency (DARPA).
\end{acks}

\bibliographystyle{ACM-Reference-Format}
\bibliography{references} 


\begin{thebibliography}{34}


\ifx \showCODEN    \undefined \def \showCODEN     #1{\unskip}     \fi
\ifx \showDOI      \undefined \def \showDOI       #1{#1}\fi
\ifx \showISBNx    \undefined \def \showISBNx     #1{\unskip}     \fi
\ifx \showISBNxiii \undefined \def \showISBNxiii  #1{\unskip}     \fi
\ifx \showISSN     \undefined \def \showISSN      #1{\unskip}     \fi
\ifx \showLCCN     \undefined \def \showLCCN      #1{\unskip}     \fi
\ifx \shownote     \undefined \def \shownote      #1{#1}          \fi
\ifx \showarticletitle \undefined \def \showarticletitle #1{#1}   \fi
\ifx \showURL      \undefined \def \showURL       {\relax}        \fi
\providecommand\bibfield[2]{#2}
\providecommand\bibinfo[2]{#2}
\providecommand\natexlab[1]{#1}
\providecommand\showeprint[2][]{arXiv:#2}

\bibitem[\protect\citeauthoryear{Alvernaz and Togelius}{Alvernaz and
  Togelius}{2017}]%
        {alvernaz2017autoencoder}
\bibfield{author}{\bibinfo{person}{Samuel Alvernaz} {and}
  \bibinfo{person}{Julian Togelius}.} \bibinfo{year}{2017}\natexlab{}.
\newblock \showarticletitle{Autoencoder-augmented neuroevolution for visual
  doom playing}. In \bibinfo{booktitle}{{\em 2017 IEEE Conference on
  Computational Intelligence and Games (CIG)}}. IEEE, \bibinfo{pages}{1--8}.
\newblock


\bibitem[\protect\citeauthoryear{Bengio, Bengio, Cloutier, and Gecsei}{Bengio
  et~al\mbox{.}}{1992}]%
        {bengio1992optimization}
\bibfield{author}{\bibinfo{person}{Samy Bengio}, \bibinfo{person}{Yoshua
  Bengio}, \bibinfo{person}{Jocelyn Cloutier}, {and} \bibinfo{person}{Jan
  Gecsei}.} \bibinfo{year}{1992}\natexlab{}.
\newblock \showarticletitle{On the optimization of a synaptic learning rule}.
  In \bibinfo{booktitle}{{\em Preprints Conf. Optimality in Artificial and
  Biological Neural Networks}}. Univ. of Texas, \bibinfo{pages}{6--8}.
\newblock


\bibitem[\protect\citeauthoryear{Blynel and Floreano}{Blynel and
  Floreano}{2002}]%
        {blynel2002levels}
\bibfield{author}{\bibinfo{person}{Jesper Blynel} {and} \bibinfo{person}{Dario
  Floreano}.} \bibinfo{year}{2002}\natexlab{}.
\newblock \showarticletitle{Levels of dynamics and adaptive behavior in
  evolutionary neural controllers}. In \bibinfo{booktitle}{{\em Proceedings of
  the seventh international conference on simulation of adaptive behavior on
  From animals to animats}}. MIT Press, \bibinfo{pages}{272--281}.
\newblock


\bibitem[\protect\citeauthoryear{Duan, Chen, Houthooft, Schulman, and
  Abbeel}{Duan et~al\mbox{.}}{2016a}]%
        {duan2016benchmarking}
\bibfield{author}{\bibinfo{person}{Yan Duan}, \bibinfo{person}{Xi Chen},
  \bibinfo{person}{Rein Houthooft}, \bibinfo{person}{John Schulman}, {and}
  \bibinfo{person}{Pieter Abbeel}.} \bibinfo{year}{2016}\natexlab{a}.
\newblock \showarticletitle{Benchmarking deep reinforcement learning for
  continuous control}. In \bibinfo{booktitle}{{\em International Conference on
  Machine Learning}}. \bibinfo{pages}{1329--1338}.
\newblock


\bibitem[\protect\citeauthoryear{Duan, Schulman, Chen, Bartlett, Sutskever, and
  Abbeel}{Duan et~al\mbox{.}}{2016b}]%
        {duan2016rl}
\bibfield{author}{\bibinfo{person}{Yan Duan}, \bibinfo{person}{John Schulman},
  \bibinfo{person}{Xi Chen}, \bibinfo{person}{Peter~L Bartlett},
  \bibinfo{person}{Ilya Sutskever}, {and} \bibinfo{person}{Pieter Abbeel}.}
  \bibinfo{year}{2016}\natexlab{b}.
\newblock \showarticletitle{RL$^2$: Fast reinforcement learning via slow
  reinforcement learning}.
\newblock \bibinfo{journal}{{\em arXiv preprint arXiv:1611.02779\/}}
  (\bibinfo{year}{2016}).
\newblock


\bibitem[\protect\citeauthoryear{Finn, Abbeel, and Levine}{Finn
  et~al\mbox{.}}{2017}]%
        {finn2017model}
\bibfield{author}{\bibinfo{person}{Chelsea Finn}, \bibinfo{person}{Pieter
  Abbeel}, {and} \bibinfo{person}{Sergey Levine}.}
  \bibinfo{year}{2017}\natexlab{}.
\newblock \showarticletitle{Model-agnostic meta-learning for fast adaptation of
  deep networks}. In \bibinfo{booktitle}{{\em Proceedings of the 34th
  International Conference on Machine Learning-Volume 70}}. JMLR. org,
  \bibinfo{pages}{1126--1135}.
\newblock


\bibitem[\protect\citeauthoryear{Floreano and Keller}{Floreano and
  Keller}{2010}]%
        {floreano2010evolution}
\bibfield{author}{\bibinfo{person}{Dario Floreano} {and}
  \bibinfo{person}{Laurent Keller}.} \bibinfo{year}{2010}\natexlab{}.
\newblock \showarticletitle{Evolution of adaptive behaviour in robots by means
  of Darwinian selection}.
\newblock \bibinfo{journal}{{\em PLoS Biol\/}} \bibinfo{volume}{8},
  \bibinfo{number}{1} (\bibinfo{year}{2010}), \bibinfo{pages}{e1000292}.
\newblock


\bibitem[\protect\citeauthoryear{Floreano and Mondada}{Floreano and
  Mondada}{1996}]%
        {floreano1996evolution}
\bibfield{author}{\bibinfo{person}{Dario Floreano} {and}
  \bibinfo{person}{Francesco Mondada}.} \bibinfo{year}{1996}\natexlab{}.
\newblock \showarticletitle{Evolution of plastic neurocontrollers for situated
  agents}. In \bibinfo{booktitle}{{\em Proc. of The Fourth International
  Conference on Simulation of Adaptive Behavior (SAB), From Animals to
  Animats}}. ETH Z{\"u}rich.
\newblock


\bibitem[\protect\citeauthoryear{Ha and Schmidhuber}{Ha and
  Schmidhuber}{2018}]%
        {ha2018recurrent}
\bibfield{author}{\bibinfo{person}{David Ha} {and} \bibinfo{person}{J{\"u}rgen
  Schmidhuber}.} \bibinfo{year}{2018}\natexlab{}.
\newblock \showarticletitle{Recurrent world models facilitate policy
  evolution}. In \bibinfo{booktitle}{{\em Advances in Neural Information
  Processing Systems}}. \bibinfo{pages}{2450--2462}.
\newblock


\bibitem[\protect\citeauthoryear{Hochreiter, Younger, and Conwell}{Hochreiter
  et~al\mbox{.}}{2001}]%
        {hochreiter2001learning}
\bibfield{author}{\bibinfo{person}{Sepp Hochreiter}, \bibinfo{person}{A~Steven
  Younger}, {and} \bibinfo{person}{Peter~R Conwell}.}
  \bibinfo{year}{2001}\natexlab{}.
\newblock \showarticletitle{Learning to learn using gradient descent}. In
  \bibinfo{booktitle}{{\em International Conference on Artificial Neural
  Networks}}. Springer, \bibinfo{pages}{87--94}.
\newblock


\bibitem[\protect\citeauthoryear{Johnson, Hofmann, Hutton, and Bignell}{Johnson
  et~al\mbox{.}}{2016}]%
        {johnson2016malmo}
\bibfield{author}{\bibinfo{person}{Matthew Johnson}, \bibinfo{person}{Katja
  Hofmann}, \bibinfo{person}{Tim Hutton}, {and} \bibinfo{person}{David
  Bignell}.} \bibinfo{year}{2016}\natexlab{}.
\newblock \showarticletitle{The Malmo Platform for Artificial Intelligence
  Experimentation.}. In \bibinfo{booktitle}{{\em IJCAI}}.
  \bibinfo{pages}{4246--4247}.
\newblock


\bibitem[\protect\citeauthoryear{Lillicrap, Hunt, Pritzel, Heess, Erez, Tassa,
  Silver, and Wierstra}{Lillicrap et~al\mbox{.}}{2015}]%
        {lillicrap2015continuous}
\bibfield{author}{\bibinfo{person}{Timothy~P Lillicrap},
  \bibinfo{person}{Jonathan~J Hunt}, \bibinfo{person}{Alexander Pritzel},
  \bibinfo{person}{Nicolas Heess}, \bibinfo{person}{Tom Erez},
  \bibinfo{person}{Yuval Tassa}, \bibinfo{person}{David Silver}, {and}
  \bibinfo{person}{Daan Wierstra}.} \bibinfo{year}{2015}\natexlab{}.
\newblock \showarticletitle{Continuous control with deep reinforcement
  learning}.
\newblock \bibinfo{journal}{{\em arXiv preprint arXiv:1509.02971\/}}
  (\bibinfo{year}{2015}).
\newblock


\bibitem[\protect\citeauthoryear{McHale and Husbands}{McHale and
  Husbands}{2004}]%
        {mchale2004gasnets}
\bibfield{author}{\bibinfo{person}{Gary McHale} {and} \bibinfo{person}{Phil
  Husbands}.} \bibinfo{year}{2004}\natexlab{}.
\newblock \showarticletitle{Gasnets and other evolvable neural networks applied
  to bipedal locomotion}.
\newblock \bibinfo{journal}{{\em From Animals to Animats\/}}
  \bibinfo{volume}{8} (\bibinfo{year}{2004}), \bibinfo{pages}{163--172}.
\newblock


\bibitem[\protect\citeauthoryear{Mishra, Rohaninejad, Chen, and Abbeel}{Mishra
  et~al\mbox{.}}{2018}]%
        {mishra2018a}
\bibfield{author}{\bibinfo{person}{Nikhil Mishra}, \bibinfo{person}{Mostafa
  Rohaninejad}, \bibinfo{person}{Xi Chen}, {and} \bibinfo{person}{Pieter
  Abbeel}.} \bibinfo{year}{2018}\natexlab{}.
\newblock \showarticletitle{A Simple Neural Attentive Meta-Learner}. In
  \bibinfo{booktitle}{{\em International Conference on Learning
  Representations}}.
\newblock
\showURL{%
\url{https://openreview.net/forum?id=B1DmUzWAW}}


\bibitem[\protect\citeauthoryear{Mnih, Kavukcuoglu, Silver, Rusu, Veness,
  Bellemare, Graves, Riedmiller, Fidjeland, Ostrovski, et~al\mbox{.}}{Mnih
  et~al\mbox{.}}{2015}]%
        {mnih2015human}
\bibfield{author}{\bibinfo{person}{Volodymyr Mnih}, \bibinfo{person}{Koray
  Kavukcuoglu}, \bibinfo{person}{David Silver}, \bibinfo{person}{Andrei~A
  Rusu}, \bibinfo{person}{Joel Veness}, \bibinfo{person}{Marc~G Bellemare},
  \bibinfo{person}{Alex Graves}, \bibinfo{person}{Martin Riedmiller},
  \bibinfo{person}{Andreas~K Fidjeland}, \bibinfo{person}{Georg Ostrovski},
  {et~al\mbox{.}}} \bibinfo{year}{2015}\natexlab{}.
\newblock \showarticletitle{Human-level control through deep reinforcement
  learning}.
\newblock \bibinfo{journal}{{\em Nature\/}} \bibinfo{volume}{518},
  \bibinfo{number}{7540} (\bibinfo{year}{2015}), \bibinfo{pages}{529}.
\newblock


\bibitem[\protect\citeauthoryear{Poulsen, Thorhauge, Funch, and Risi}{Poulsen
  et~al\mbox{.}}{2017}]%
        {poulsen2017dlne}
\bibfield{author}{\bibinfo{person}{Andreas~Precht Poulsen},
  \bibinfo{person}{Mark Thorhauge}, \bibinfo{person}{Mikkel~Hvilshj Funch},
  {and} \bibinfo{person}{Sebastian Risi}.} \bibinfo{year}{2017}\natexlab{}.
\newblock \showarticletitle{DLNE: A hybridization of deep learning and
  neuroevolution for visual control}. In \bibinfo{booktitle}{{\em 2017 IEEE
  Conference on Computational Intelligence and Games (CIG)}}. IEEE,
  \bibinfo{pages}{256--263}.
\newblock


\bibitem[\protect\citeauthoryear{Rakelly, Zhou, Finn, Levine, and
  Quillen}{Rakelly et~al\mbox{.}}{2019}]%
        {rakelly2019efficient}
\bibfield{author}{\bibinfo{person}{Kate Rakelly}, \bibinfo{person}{Aurick
  Zhou}, \bibinfo{person}{Chelsea Finn}, \bibinfo{person}{Sergey Levine}, {and}
  \bibinfo{person}{Deirdre Quillen}.} \bibinfo{year}{2019}\natexlab{}.
\newblock \showarticletitle{Efficient Off-Policy Meta-Reinforcement Learning
  via Probabilistic Context Variables}. In \bibinfo{booktitle}{{\em
  International Conference on Machine Learning}}. \bibinfo{pages}{5331--5340}.
\newblock


\bibitem[\protect\citeauthoryear{Risi and Stanley}{Risi and Stanley}{2019a}]%
        {risi2019deep}
\bibfield{author}{\bibinfo{person}{Sebastian Risi} {and}
  \bibinfo{person}{Kenneth~O Stanley}.} \bibinfo{year}{2019}\natexlab{a}.
\newblock \showarticletitle{Deep neuroevolution of recurrent and discrete world
  models}. In \bibinfo{booktitle}{{\em Proceedings of the Genetic and
  Evolutionary Computation Conference}}. \bibinfo{pages}{456--462}.
\newblock


\bibitem[\protect\citeauthoryear{Risi and Stanley}{Risi and Stanley}{2019b}]%
        {risi2019improving}
\bibfield{author}{\bibinfo{person}{Sebastian Risi} {and}
  \bibinfo{person}{Kenneth~O Stanley}.} \bibinfo{year}{2019}\natexlab{b}.
\newblock \showarticletitle{Improving Deep Neuroevolution via Deep Innovation
  Protection}.
\newblock \bibinfo{journal}{{\em arXiv preprint arXiv:2001.01683\/}}
  (\bibinfo{year}{2019}).
\newblock


\bibitem[\protect\citeauthoryear{Rothfuss, Lee, Clavera, Asfour, and
  Abbeel}{Rothfuss et~al\mbox{.}}{2019}]%
        {rothfuss2018promp}
\bibfield{author}{\bibinfo{person}{Jonas Rothfuss}, \bibinfo{person}{Dennis
  Lee}, \bibinfo{person}{Ignasi Clavera}, \bibinfo{person}{Tamim Asfour}, {and}
  \bibinfo{person}{Pieter Abbeel}.} \bibinfo{year}{2019}\natexlab{}.
\newblock \showarticletitle{Pro{MP}: Proximal Meta-Policy Search}. In
  \bibinfo{booktitle}{{\em International Conference on Learning
  Representations}}.
\newblock
\showURL{%
\url{https://openreview.net/forum?id=SkxXCi0qFX}}


\bibitem[\protect\citeauthoryear{Rumelhart, Hinton, Williams,
  et~al\mbox{.}}{Rumelhart et~al\mbox{.}}{1988}]%
        {rumelhart1988learning}
\bibfield{author}{\bibinfo{person}{David~E Rumelhart},
  \bibinfo{person}{Geoffrey~E Hinton}, \bibinfo{person}{Ronald~J Williams},
  {et~al\mbox{.}}} \bibinfo{year}{1988}\natexlab{}.
\newblock \showarticletitle{Learning representations by back-propagating
  errors}.
\newblock \bibinfo{journal}{{\em Cognitive modeling\/}} \bibinfo{volume}{5},
  \bibinfo{number}{3} (\bibinfo{year}{1988}), \bibinfo{pages}{1}.
\newblock


\bibitem[\protect\citeauthoryear{Salimans, Ho, Chen, Sidor, and
  Sutskever}{Salimans et~al\mbox{.}}{2017}]%
        {salimans2017evolution}
\bibfield{author}{\bibinfo{person}{Tim Salimans}, \bibinfo{person}{Jonathan
  Ho}, \bibinfo{person}{Xi Chen}, \bibinfo{person}{Szymon Sidor}, {and}
  \bibinfo{person}{Ilya Sutskever}.} \bibinfo{year}{2017}\natexlab{}.
\newblock \showarticletitle{Evolution strategies as a scalable alternative to
  reinforcement learning}.
\newblock \bibinfo{journal}{{\em arXiv preprint arXiv:1703.03864\/}}
  (\bibinfo{year}{2017}).
\newblock


\bibitem[\protect\citeauthoryear{Schmidhuber, Zhao, and Wiering}{Schmidhuber
  et~al\mbox{.}}{1996}]%
        {schmidhuber1996simple}
\bibfield{author}{\bibinfo{person}{Juergen Schmidhuber}, \bibinfo{person}{Jieyu
  Zhao}, {and} \bibinfo{person}{MA Wiering}.} \bibinfo{year}{1996}\natexlab{}.
\newblock \showarticletitle{Simple principles of metalearning}.
\newblock \bibinfo{journal}{{\em Technical report IDSIA\/}}
  \bibinfo{volume}{69} (\bibinfo{year}{1996}), \bibinfo{pages}{1--23}.
\newblock


\bibitem[\protect\citeauthoryear{Schweighofer and Doya}{Schweighofer and
  Doya}{2003}]%
        {schweighofer2003meta}
\bibfield{author}{\bibinfo{person}{Nicolas Schweighofer} {and}
  \bibinfo{person}{Kenji Doya}.} \bibinfo{year}{2003}\natexlab{}.
\newblock \showarticletitle{Meta-learning in reinforcement learning}.
\newblock \bibinfo{journal}{{\em Neural Networks\/}} \bibinfo{volume}{16},
  \bibinfo{number}{1} (\bibinfo{year}{2003}), \bibinfo{pages}{5--9}.
\newblock


\bibitem[\protect\citeauthoryear{Soltoggio, Bullinaria, Mattiussi, D{\"u}rr,
  and Floreano}{Soltoggio et~al\mbox{.}}{2008}]%
        {soltoggio2008evolutionary}
\bibfield{author}{\bibinfo{person}{Andrea Soltoggio}, \bibinfo{person}{John~A
  Bullinaria}, \bibinfo{person}{Claudio Mattiussi}, \bibinfo{person}{Peter
  D{\"u}rr}, {and} \bibinfo{person}{Dario Floreano}.}
  \bibinfo{year}{2008}\natexlab{}.
\newblock \showarticletitle{Evolutionary advantages of neuromodulated
  plasticity in dynamic, reward-based scenarios}. In \bibinfo{booktitle}{{\em
  Proceedings of the 11th international conference on artificial life (Alife
  XI)}}. MIT Press, \bibinfo{pages}{569--576}.
\newblock


\bibitem[\protect\citeauthoryear{Soltoggio, Stanley, and Risi}{Soltoggio
  et~al\mbox{.}}{2018}]%
        {soltoggio2018born}
\bibfield{author}{\bibinfo{person}{Andrea Soltoggio},
  \bibinfo{person}{Kenneth~O Stanley}, {and} \bibinfo{person}{Sebastian Risi}.}
  \bibinfo{year}{2018}\natexlab{}.
\newblock \showarticletitle{Born to learn: the inspiration, progress, and
  future of evolved plastic artificial neural networks}.
\newblock \bibinfo{journal}{{\em Neural Networks\/}} (\bibinfo{year}{2018}).
\newblock


\bibitem[\protect\citeauthoryear{Stanley, Clune, Lehman, and
  Miikkulainen}{Stanley et~al\mbox{.}}{2019}]%
        {stanley2019designing}
\bibfield{author}{\bibinfo{person}{Kenneth~O Stanley}, \bibinfo{person}{Jeff
  Clune}, \bibinfo{person}{Joel Lehman}, {and} \bibinfo{person}{Risto
  Miikkulainen}.} \bibinfo{year}{2019}\natexlab{}.
\newblock \showarticletitle{Designing neural networks through neuroevolution}.
\newblock \bibinfo{journal}{{\em Nature Machine Intelligence\/}}
  \bibinfo{volume}{1}, \bibinfo{number}{1} (\bibinfo{year}{2019}),
  \bibinfo{pages}{24--35}.
\newblock


\bibitem[\protect\citeauthoryear{Stanley and Miikkulainen}{Stanley and
  Miikkulainen}{2002}]%
        {stanley2002evolving}
\bibfield{author}{\bibinfo{person}{Kenneth~O Stanley} {and}
  \bibinfo{person}{Risto Miikkulainen}.} \bibinfo{year}{2002}\natexlab{}.
\newblock \showarticletitle{Evolving neural networks through augmenting
  topologies}.
\newblock \bibinfo{journal}{{\em Evolutionary computation\/}}
  \bibinfo{volume}{10}, \bibinfo{number}{2} (\bibinfo{year}{2002}),
  \bibinfo{pages}{99--127}.
\newblock


\bibitem[\protect\citeauthoryear{Such, Madhavan, Conti, Lehman, Stanley, and
  Clune}{Such et~al\mbox{.}}{2017}]%
        {such2017deep}
\bibfield{author}{\bibinfo{person}{Felipe~Petroski Such},
  \bibinfo{person}{Vashisht Madhavan}, \bibinfo{person}{Edoardo Conti},
  \bibinfo{person}{Joel Lehman}, \bibinfo{person}{Kenneth~O Stanley}, {and}
  \bibinfo{person}{Jeff Clune}.} \bibinfo{year}{2017}\natexlab{}.
\newblock \showarticletitle{Deep neuroevolution: Genetic algorithms are a
  competitive alternative for training deep neural networks for reinforcement
  learning}.
\newblock \bibinfo{journal}{{\em arXiv preprint arXiv:1712.06567\/}}
  (\bibinfo{year}{2017}).
\newblock


\bibitem[\protect\citeauthoryear{Thrun and Pratt}{Thrun and Pratt}{1998}]%
        {thrun1998learning}
\bibfield{author}{\bibinfo{person}{Sebastian Thrun} {and}
  \bibinfo{person}{Lorien Pratt}.} \bibinfo{year}{1998}\natexlab{}.
\newblock \showarticletitle{Learning to learn: Introduction and overview}.
\newblock In \bibinfo{booktitle}{{\em Learning to learn}}.
  \bibinfo{publisher}{Springer}, \bibinfo{pages}{3--17}.
\newblock


\bibitem[\protect\citeauthoryear{Wang, Kurth-Nelson, Tirumala, Soyer, Leibo,
  Munos, Blundell, Kumaran, and Botvinick}{Wang et~al\mbox{.}}{2016}]%
        {wang1611learning}
\bibfield{author}{\bibinfo{person}{Jane~X Wang}, \bibinfo{person}{Zeb
  Kurth-Nelson}, \bibinfo{person}{Dhruva Tirumala}, \bibinfo{person}{Hubert
  Soyer}, \bibinfo{person}{Joel~Z Leibo}, \bibinfo{person}{R{\'e}mi Munos},
  \bibinfo{person}{Charles Blundell}, \bibinfo{person}{Dharshan Kumaran}, {and}
  \bibinfo{person}{Matt Botvinick}.} \bibinfo{year}{2016}\natexlab{}.
\newblock \showarticletitle{Learning to reinforcement learn, 2016}.
\newblock \bibinfo{journal}{{\em arXiv preprint arXiv:1611.05763\/}}
  (\bibinfo{year}{2016}).
\newblock


\bibitem[\protect\citeauthoryear{Werbos}{Werbos}{1982}]%
        {werbos1982applications}
\bibfield{author}{\bibinfo{person}{Paul~J Werbos}.}
  \bibinfo{year}{1982}\natexlab{}.
\newblock \showarticletitle{Applications of advances in nonlinear sensitivity
  analysis}.
\newblock In \bibinfo{booktitle}{{\em System modeling and optimization}}.
  \bibinfo{publisher}{Springer}, \bibinfo{pages}{762--770}.
\newblock


\bibitem[\protect\citeauthoryear{Yao}{Yao}{1999}]%
        {yao1999evolving}
\bibfield{author}{\bibinfo{person}{Xin Yao}.} \bibinfo{year}{1999}\natexlab{}.
\newblock \showarticletitle{Evolving artificial neural networks}.
\newblock \bibinfo{journal}{{\it Proc. IEEE}} \bibinfo{volume}{87},
  \bibinfo{number}{9} (\bibinfo{year}{1999}), \bibinfo{pages}{1423--1447}.
\newblock


\bibitem[\protect\citeauthoryear{Zintgraf, Shiarli, Kurin, Hofmann, and
  Whiteson}{Zintgraf et~al\mbox{.}}{2019}]%
        {zintgraf2019fast}
\bibfield{author}{\bibinfo{person}{Luisa Zintgraf}, \bibinfo{person}{Kyriacos
  Shiarli}, \bibinfo{person}{Vitaly Kurin}, \bibinfo{person}{Katja Hofmann},
  {and} \bibinfo{person}{Shimon Whiteson}.} \bibinfo{year}{2019}\natexlab{}.
\newblock \showarticletitle{Fast Context Adaptation via Meta-Learning}. In
  \bibinfo{booktitle}{{\em International Conference on Machine Learning}}.
  \bibinfo{pages}{7693--7702}.
\newblock


\end{thebibliography}

\end{document}